\documentclass{IJCAS}

\usepackage{url}
\usepackage{array,tabularx}
\usepackage{threeparttable}
\usepackage{multicol} 
\usepackage{multirow}
\usepackage{comment}
\usepackage[dvipsnames]{xcolor}

\journalvolumn{VV}
\journalnumber{X}
\journalyear{YYYY}
\setarticlestartpagenumber{1}

\allowdisplaybreaks

\begin{document}
\title{Efficient Computation of Whole-Body Control Utilizing Simplified Whole-Body Dynamics via Centroidal Dynamics}

\author{Junewhee Ahn\orcid{}, Jaesug Jung\orcid{}, Yisoo Lee\orcid{}, Hokyun Lee\orcid{}, Sami Haddadin\orcid{} and Jaeheung Park*\orcid{}}

\begin{abstract}
In this study, we present a novel method for enhancing the computational efficiency of whole-body control for humanoid robots, a challenge accentuated by their high degrees of freedom. 
The reduced-dimension rigid body dynamics of a floating base robot is constructed by segmenting its kinematic chain into constrained and unconstrained chains, simplifying the dynamics of the unconstrained chain through the centroidal dynamics.
The proposed dynamics model is possible to be applied to whole-body control methods, allowing the problem to be divided into two parts for more efficient computation.
The efficiency of the framework is demonstrated by comparative experiments in simulations.
The calculation results demonstrate a significant reduction in processing time, highlighting an improvement over the times reported in current methodologies. 
Additionally, the results also shows the computational efficiency increases as the degrees of freedom of robot model increases.
\end{abstract}


\maketitle

\makeAuthorInformation{
Manuscript received January 10, 2021; revised March 10, 2021; accepted May 10, 2021. Recommended by Associate Editor Soon-Shin Lee under the direction of Editor Milton John. This work was supported by the National Research Foundation of Korea (NRF) grant funded by the Korea government (MSIT) (No. 2021R1A2C3005914). It was also supported by TUM AGENDA 2030, funded by the Federal Ministry of Education and Research (BMBF) and the Free State of Bavaria under the Excellence Strategy of the Federal Government and the Länder as well as by the Hightech Agenda Bavaria.\\

J. Ahn, H. Lee and J. Park are with the Graduate School of Convergence Science and Technology, Seoul National University (SNU), Gwanak-ro 1, Gwanak-gu, Seoul, Republic of Korea. (e-mails: \{june992, hkleetony, park73\}@snu.ac.kr).
J. Jung is with Global Infrastructure and manufacturing, Samsung Electronics. (e-mails: jays.jung@samsung.com).
Y. Lee is with the Korea Institute of Science and Technology (KIST), Seoul, 02792, Republic of Korea (e-mails: yisoo.lee@kist.re.kr).
S. Haddadin is with Technical University of Munich, Germany; TUM School of Computation, Information and Technology (CIT); Chair of Robotics and Systems Intelligence (RSI); Munich Institute of Robotics and Machine Intelligence (MIRMI) (e-mails: haddadin@tum.de).
J. Park is also with ASRI, RICS, Seoul National University, Republic of Korea, and Advanced Institutes and Advanced Institutes of Convergence Technology(AICT).

* Corresponding author.
}

\runningtitle{2022}{Junewhee Ahn, Jaesug Jung, Yisoo Lee, Hokyun Lee, Sami Haddadin and Jaeheung Park}{Manuscript Template for the International Journal of Control, Automation, and Systems: ICROS {\&} KIEE}{xxx}{xxxx}{x}

\section{INTRODUCTION}

The field of humanoid robotics has witnessed significant advances in recent years, driven by the demand for robots that can operate in environments designed for humans and perform tasks ranging from simple repetitive actions to complex, dynamic interactions. 
One of the essential components in realizing this is the development of Whole-Body Control (WBC) frameworks, initially explored in foundational research \cite{park2006contact, sentis2006whole}.
WBC facilitates the execution of multi-task control at the torque level within the operational space, carefully accounting for contact dynamics and ensuring consistency with various constraints. 

Recent studies on whole-body control methods for walking robots have predominantly focused on optimization-based techniques, demonstrating significant improvements in control performance. One prominent method in this domain is Model Predictive Control (MPC), which predicts future states and optimizes trajectory accordingly. 
MPC has been implemented in various ways for humanoid robots, primarily in two major approaches.
The first approach employs a whole-body model. 
Although substantial research has been conducted using whole-body models \cite{koenemann2015whole, neunert2018whole}, the complexity of these methods poses a major challenge in reducing computational burden \cite{katayama2023model, wensing2023optimization}.
The second approach leverages simplified dynamic models such as centroidal dynamics~\cite{orin2013centroidal}, Linear inverted pendulum model~\cite{kajita20013d}, and Single Rigid Body Dynamics. 
Several research on MPC \cite{rathod2021model, jeon2022online} uses the simplified models and then extends the optimized trajectory of simplfied model to the whole-body control. 
However, these simplified models have clear limitations, particularly in directly considering crucial contact constraints for walking robots.

The method of computing whole-body control inputs from the optimized trajectory of simplified model is known as instantaneous WBC (iWBC). Since the trajectory optimizations like MPC require significant computation time, it is crucial to reduce both the computation time and latency in iWBC to enhance performance.
This latency is not merely a delay; it fundamentally undermines control stability, which is a critical concern highlighted in \cite{stepan2006stability, jung2017analysis}.

Approaches to iWBC through optimization started with Quadratic Programming (QP), and the QP-based WBC formulation \cite{collette2008robust} takes into account robot dynamics and kinematics by considering multiple constraints and other practical conditions \cite{qpfeng2015optimization,qpkuindersma2014efficiently,qpmesesan2019dynamic}.
Subsequently, several studies \cite{de2009prioritized,herzog2016momentum} utilized the Hierarchical Quadratic Programming (HQP), also known as Lexicographic QP (LQP), to maintain the explicit hierarchy of tasks and constraints. To reduce the computational burden of HQP, several methods have been proposed:
A research \cite{herzog2016momentum} reduced computational cost by decomposing the optimization variables, especially the constraint by whole-body dynamics. 
Then \cite{de2010feature} focused on developing solvers to reduce the computational load of HQP. 
Further, dedicated solvers \cite{escande2014hierarchical}, \cite{dimitrov2015efficient} for HQP were developed, achieving significant computational improvements.

Despite these advancements in solvers, many optimization-based control studies continue to rely on off-the-shelf optimization solvers. 
Solvers such as OSQP \cite{osqp}, qpOASES \cite{Ferreau2014}, acados \cite{acados}, and quadprog are frequently used in research \cite{rathod2021model, jeon2022online, raiola2020simple, klemm2020lqr, ramuzat2021comparison} due to the complexity of developing custom solvers. These pre-built solvers facilitate optimization tasks by providing robust and efficient solutions, allowing researchers to focus on other aspects of control system design and implementation.
From this perspective, the study \cite{sathya2021weighted} also formulated a lexicographic problem using the weighted method, demonstrating that significant performance improvements can be achieved with off-the-shelf solvers, even when compared with dedicated solvers.

Unlike conventional approaches to iWBC, a novel approach has been proposed in  \cite{leeoswbc, yslee}, simplifying the constraint itself through operational space formulation.
This approach resulted in a greater reduction in optimization variables and constraint dimension to previous methods \cite{herzog2016momentum}, thus significantly enhancing computational performance even with the off-the-shelf optimization problem solvers.
However, these methods not only consume additional computation time to construct the Operational Space Formulation (OSF) but also require a thorough understanding of the OSF. Moreover, inequality constraints in the joint space are considered indirectly through the operational space, which can potentially result in sub-optimal solutions.

This paper proposes a method that utilizes the projection into a specific space to split the dynamics model into two parts.
The proposed method then solve the iWBC problem sequentially using the split models.
The first part of the dynamics model proposed in this paper simplifies the joints that are not used to describe the constrained link, which typically associated with contact constraint. 
The dynamics of these unconstrained joints are summarized through centroidal dynamics, with a focus on environmental interaction, which is crucial for humanoid robots. 
For the second part, the dynamics of unconstrained link's joint space are considered. This two-part configuration allows the entire problem to be divided and addressed as two separate problems.
The proposed method leverages the advantages of both whole-body dynamics and centroidal dynamics, offering a comprehensive approach to efficiently manage the computation problem.

The proposed method's approach is different from conventional approaches of iWBC by altering the fundamental dynamics model itself.
The advantage of the proposed approach is that it allows the direct application of conventional methods, including solver-based methods, constraint handling, and optimization construction approaches, to our method.

The contributions of this study, which introduces a novel framework for improving efficiency, are as follows:
\begin{itemize}
    \item Introduces a novel method for constructing reduced dimension rigid body dynamics in WBC, achieving high computational efficiency.
    \item Demonstrates how the reduced dynamics model can be integrated into various formulations of iWBC, providing a distinct approach that differentiates it from conventional methods to iWBC.
\end{itemize}
To elucidate these contributions, the paper is structured as follows: Section~\ref{section2:wholenbodycontrol} provides an overview of whole-body control. Section~\ref{section3:constructionofucdyn} introduces a method for constructing reduced dynamics. Section~\ref{section4:problemformulation} discusses the application of the proposed method to conventional approaches. Section~\ref{setion5:evaluation} presents the results. Finally, Section~\ref{section6:conclusion} concludes the paper with a discussion based on the validation findings.

\section{CONSTRAINT FOR WHOLE-BODY CONTROL}\label{section2:wholenbodycontrol}
Whole-body dynamics is a fundamental concept in robotics that expresses the dynamic behavior of a floating-base robot composed of multiple rigid bodies connected by joints. It describes how internal and external forces affect the robot's movements and can be expressed by the following equation:
\begin{equation}
    \mathbf{M}(\mathbf{q})\mathbf{\ddot{q}} + \mathbf{b}(\mathbf{q},\mathbf{\dot{q}}) + \mathbf{g}(\mathbf{q}) +\mathbf{J}_c(\mathbf{q})^T \mathbf{F}_c = \mathbf{S}^T \boldsymbol{\Gamma},\label{eq:dynamics}
\end{equation}
where $\mathbf{q}\in\mathbb{R}^{n+6}$ is the generalized joint position vector of the system, and $n$ is the degrees of freedom (DOF) of the robot actuated joints. $\mathbf{M}(\mathbf{q})\in\mathbb{R}^{(n+6) \times (n+6)}$ is the inertia matrix, $\mathbf{b}(\mathbf{q},\mathbf{\dot{q}})\in\mathbb{R}^{n+6}$ is the Coriolis and centrifugal force vector, $\mathbf{g}(\mathbf{q})\in\mathbb{R}^{n+6}$ is the gravity force vector, $\mathbf{S}^T\in\mathbb{R}^{{(n+6)}\times n}$ is a selection matrix that specifies actuated joints from system joints, $\mathbf{J}_c(\mathbf{q})\in\mathbb{R}^{c\times (n+6)}$ is the constraint Jacobian, which usually used to describe the contact constraint of the robot. $c$ is the constrained DOF, and $\mathbf{F}_c\in\mathbb{R}^c$ is a vector that concatenate all constraint forces.
The iWBC solves the optimization problem that considers whole-body dynamics to generate a feasible joint control solution when task space trajectories are given.
Various inequality and equality constraints must be considered for the iWBC optimization problem. 
For a humanoid robot interacting with the external environment, the necessary constraints include the equality constraint with~\eqref{eq:dynamics} to describe the robot's system dynamics. Additionally, the following inequality and equality constraints are required:
\begin{align}
&\textrm{(stationary contact) } \mathbf{J}_c \dot{\mathbf{q}} = \dot{\mathbf{J}_c}\dot{\mathbf{q}} + \mathbf{J}_c \ddot{\mathbf{q}} = 0 \label{eq:ocp1_cc}\\
&\textrm{(contact wrench cone) } \mathbf{C} \mathbf{F}_{c} \leq 0 \label{eq:ocp1_fc}\\
&\textrm{(task control) } \mathbf{J}_{task}\mathbf{\ddot{q}} + \mathbf{\dot{J}}_{task}\mathbf{\dot{q}} = \mathbf{\ddot{x}}^{ref}_{task}\label{eq:ocp1_task}\\
&\textrm{(torque limits) } \boldsymbol{\Gamma}_{min} < \boldsymbol{\Gamma} < \boldsymbol{\Gamma}_{max}, \cdots 
\end{align}
where $\mathbf{J}_{task}$ is the task Jacobian and $\mathbf{\ddot{x}}^{ref}_{task}$ is a reference task acceleration that will correspond to a desired control method (e.g. LQR, MPC, or PD-controller), and $\mathbf{C}$ is a matrix for a contact wrench cone that accounts for friction and CoP constraints. 

The system dynamics constraint in \eqref{eq:dynamics} can be decomposed as proposed in \cite{herzog2016momentum}, as follows,
\begin{align}
    \mathbf{M}_u\mathbf{\ddot{q}} + \mathbf{b}_u + \mathbf{g}_u + \mathbf{J}_{c,u}^T \mathbf{F}_{c} &= 0 \label{eq:eomfloatingbase}\\
    \mathbf{M}_l\mathbf{\ddot{q}} + \mathbf{b}_l + \mathbf{g}_l + \mathbf{J}_{c,l}^T \mathbf{F}_c &= \boldsymbol{\Gamma}\label{eq:eomjoint} 
\end{align}
Equation \eqref{eq:eomfloatingbase} represents the top six equations of \eqref{eq:dynamics}, and Equation \eqref{eq:eomjoint} comprises the bottom \(n\) equations of \eqref{eq:dynamics}. 
Equation \eqref{eq:eomfloatingbase} describes the Newton-Euler equations for the floating base and can be used as the dynamics constraint instead of \eqref{eq:dynamics}.
Through Equation \eqref{eq:eomjoint}, control torques, $\boldsymbol{\Gamma}$ can be calculated with \(\ddot{\mathbf{q}}\) and \(\mathbf{F}_c\).
Therefore, by excluding torque from the decision variable $\mathbf{x}$ the optimization problem can be solved efficiently.

\section{CONSTRUCTION OF UNCONSTRAINED IMPLICIT DYNAMICS}\label{section3:constructionofucdyn}
\subsection{Categorization and Simplification of Kinematic Chains}\label{sub1}
In the proposed method, the kinematic chain of floating-base robots is divided into three types. The \textit{virtual chain} consists of virtual joints that represent the robot's floating base relative to the inertial frame. 
The \textit{constraint chain} includes the links from the floating base to the constraint link.
The constraint links are typically described by \( \mathbf{J}_c \), forming the constrained space (usually used as the contact space for walking robot).
The \textit{unconstrained chain} comprises the links not included in the constraint chain. 
Additionally, the \textit{constraint chain} and the \textit{unconstrained chain} are composed of joints required to describe the motion of each link from the frame of the floating base. The DOF of these joints are \(n_{cc}\) and \(n_{uc}\), respectively.

\begin{figure}[t]
\centering
\includegraphics[scale=1.0]{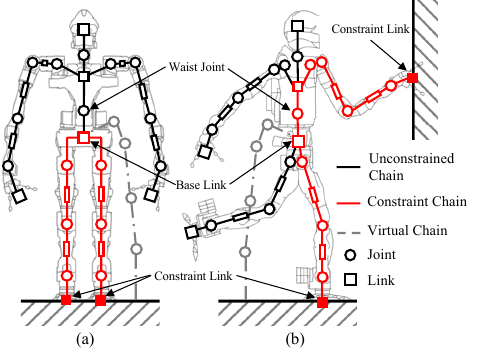}
\caption{Simplified representation of the robot's links, joints, illustrating the categorization into virtual chain, constraint chain, and unconstrained chain.}
\label{diag}
\end{figure}

Fig.~\ref{diag} illustrates this classification of kinematic chains using the simplified representation of robot. In Fig.~\ref{diag} (a), the kinematic chain of the constraint chain and the unconstrained chain are clearly separated from the floating base, making the distinction straight forward. However, in Fig.~\ref{diag} (b), the left arm and neck forming the unconstrained chain are not directly connected to the floating base. In this case, the waist joint, which is included in the constraint chain, is also required for the unconstrained chain to describe the motion from the frame fo the floating base.

\begin{figure}[t]
\centering
\includegraphics[scale=1.0]{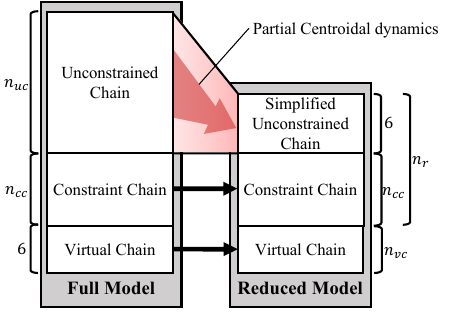}
\caption{Composition comparison between the full model and the reduced-dimension model. $n_{vc}$, $n_{cc}$, $n_{uc}$ represent the DOF of the virtual chain, constraint chain, and unconstrained chain, respectively.}
\label{figDOF}
\end{figure}

In the proposed framework, the dynamical model of the unconstrained chain with more than six DOF is simplified into six DOF utilizing the centroidal dynamics.
The relationship between the original robot model and the reduced model can be seen in Fig.~\ref{figDOF}. 
By reducing the \(n_{uc}\) DOF of the unconstrained chain to 6 DOF, the robot's model DOF can be simplified from \(n\) to \(n_{cc} + 6\).
This reduction significantly decreases computational complexity in control algorithms, especially for robots with a large number of joints.

While the kinematic chain is divided into two types based on constraints in this paper, the proposed method is not limited to this constraint specific criterion. The crucial point is that it compresses the DOF of one of the chains. Therefore, the designation of the floating base or the scope of the kinematic chain to be compressed can be adjusted according to the user's goals to enhance computational efficiency.


\subsection{Interpreting Unconstrained Chain with Centroidal dynamics}
The \textit{centroidal dynamics} \cite{orin2013centroidal} describes the total linear momentum $\mathbf{l}_G\in\mathbb{R}^3$ and the total angular momentum $\mathbf{k}_G\in\mathbb{R}^3$ of the system. The centroidal momentum is composed as $\mathbf{h}_G=[\mathbf{k}_G^T \quad \mathbf{l}_G^T]^T\in\mathbb{R}^6$.

Here, the centroidal dynamics is utilized to interpret the average movement of unconstrained chain. 
The centroidal momentum of the unconstrained chain, $\mathbf{h}_{G,uc}$ can be obtained with following, 
\begin{eqnarray}
\mathbf{h}_{G,uc} = \mathbf{M}_{G,uc}(\mathbf{q}_{uc})\dot{\mathbf{q}}_{uc}, 
\label{eq_hG}
\end{eqnarray}
where $\mathbf{M}_{G,uc}(\mathbf{q}_{uc})\in\mathbb{R}^{6\times n_{uc}}$ is the \textit{centroidal momentum matrix} \cite{orin2013centroidal} of the unconstrained chain. The centroidal momentum matrix describes the relationship between the joint coordinate and the centroidal momentum.

The average spatial velocity can be obtained with multiplying the inverse of the inertia matrix of the unconstrained chain with \eqref{eq_hG},
\begin{equation}
    \mathbf{v}_{G,uc} = \mathbf{I}_{G,uc}^{-1}\mathbf{M}_{G,uc}(\mathbf{q}_{uc})\dot{\mathbf{q}}_{uc}.\label{eq:vgnc}
\end{equation}
Where $\mathbf{I}_{G,uc} \in \mathbb{R}^{6\times 6}$ is the centroidal inertial matrix of the simplified unconstrained chain.

With \eqref{eq:vgnc}, the centroidal inertial Jacobian matrix of unconstrained chain that describes the relation between the average spatial velocity and joint space velocity can be obtained as follows:
\begin{equation}
\mathbf{J}_{G,uc} = \mathbf{I}_{G,uc}^{-1} \mathbf{M}_{G,uc}(\mathbf{q}_{uc}).
\label{eq_JG}
\end{equation}
Consequently, it is possible to interpret the average spatial movement of unconstrained chain with \eqref{eq_JG}.

\subsection{Construction of Reduced Whole-Body Dynamics}\label{sub2}

By projecting the dynamics of the unconstrained chain into a six DOF centroidal space, the joint space can be reconstructed as the reduced dimension vector ${\mathbf{q}}_{r}$.
\begin{equation}
\mathbf{\dot{q}}_{r} = {\begin{bmatrix}
    {\mathbf{\dot{q}}_{vc}^T } & {\mathbf{\dot{q}}_{cc}^T }  &{{^b}\mathbf{v}_{G,uc}^T}
    \end{bmatrix}}^T, \label{jointrelation}
\end{equation} 
where $\mathbf{\dot{q}}_{vc} \in \mathbb{R}^6$ is the generalized velocity of the virtual chain joints from an inertial frame,  $\mathbf{\dot{q}}_{cc} \in  \mathbb{R}^{ n_{cc}}$ is the generalized joint velocity of the constraint chain joints. 
The superscript preceding a variable denotes its reference frame. 
Here, $^b \mathbf{v}_{G,uc}$ indicates that \(\mathbf{v}_{G,uc}\) is the spatial velocity of unconstrained chain relative to the reference frame of the robot base.

The term $\mathbf{\dot{q}}_{r}$ can be obtained from $\mathbf{\dot{q}}$ as follows by defining the matrix $\mathbf{J}_{r} \in \mathbb{R}^{n_{r} \times (n+6)}$.
\begin{equation}\label{eqnrelation}
    \mathbf{\dot{q}}_{r} = \mathbf{J}_{r} \mathbf{\dot{q}},
\end{equation}
where
\begin{equation}
    \mathbf{J}_{r} = \begin{bmatrix}
    \mathbf{S}_{vc}^T &
    \mathbf{S}_{cc}^T & 
({^b}\mathbf{J}_{G,uc}\mathbf{S}_{uc})^T
\end{bmatrix}^T,
\end{equation}
$\mathbf{S}_{vc} \in \mathbb{R}^{6 \times (n+6)}$, $\mathbf{S}_{cc} \in \mathbb{R}^{n_{cc} \times (n+6)}$, $\mathbf{S}_{uc} \in \mathbb{R}^{n_{uc} \times (n+6)}$ are selection matrices that select virtual chain, contact chain, and unconstrained chain joints from all joints, respectively.
The centroidal inertial Jacobian matrix of the unconstrained chain ${^b}\mathbf{J}_{G,uc}\in \mathbb{R}^{6 \times n_{uc}}$ can be obtained from \eqref{eq_JG}.
The Jacobian matrix $\mathbf{J}_{r}$ define the reduced-model space as task space. 

With \eqref{eqnrelation}, the whole-body dynamics of the system \eqref{eq:dynamics} is projected into the reduced-model space,

\begin{subequations}\label{eq:dynamics_reduce}
\begin{equation}
\mathbf{M}_r\mathbf{\ddot{q}}_r + \mathbf{b}_r + \mathbf{g}_r + \mathbf{J}_{c,r}^{T}\mathbf{F}_{c} = \mathbf{S}_r^T
\boldsymbol{\Gamma}_r,
\tag{\ref{eq:dynamics_reduce}}
\end{equation}
where
\begin{align}
    \mathbf{M}_r &= ( \mathbf{J}_{r} \mathbf{M}^{-1} \mathbf{J}_{r}^{T}) ^ {-1},\\
    \mathbf{b}_r &= \mathbf{M}_r\{\mathbf{J}_{r}\mathbf{M}^{-1}\mathbf{b} - \mathbf{\dot{J}}_{r}\mathbf{\dot{q}}\},\\
    \mathbf{g}_r &= \mathbf{M}_r\mathbf{J}_{r}\mathbf{M}^{-1}\mathbf{g}, \\
    {\bar{\mathbf{J}}_{r}}^T &= \mathbf{M}_r\mathbf{J}_{r}\mathbf{M}^{-1}, \\
    \mathbf{J}_{c,r}^{T} &= {\bar{\mathbf{J}}_{r}}^T\mathbf{J}_{c}^{T}.\label{eqJcrT}
\end{align}
\end{subequations}

\noindent The matrix $\mathbf{M}_r$, $\mathbf{b}_r$, and $\mathbf{g}_r$ are the projections of $\mathbf{M}$, $\mathbf{b}$, and $\mathbf{g}$ into the reduced model space, respectively.
${\bar{\mathbf{J}}_{r}}^T$ denotes the dynamically consistent inverse of $\mathbf{J}_{r}$.
$\mathbf{S}_r^T\in\mathbb{R}^{(n_r+6) \times n_r}$ is a selection matrix that specifies actuated joint for reduced model dynamics.
$\mathbf{J}_{c,r}$ is the reduced-model space projection of contact Jacobian.
Here, $\mathbf{J}_{c,r}^{T}$ can be more simply computed as
\begin{equation}
    \mathbf{J}_{c,r}^T = \begin{bmatrix}
        \mathbf{S}_{vc}^T &
        \mathbf{S}_{cc}^T &
        (\mathbf{0}_{6\times (n+6)})^T
    \end{bmatrix}^T \mathbf{J}_{c}^T,
\end{equation}
instead of \eqref{eqJcrT} since the contact Jacobian matrix inherently does not include components from the unconstrained chain.
The vector $\boldsymbol{\Gamma}_r\in\mathbb{R}^{n_r}$ represents the actuated joint torques of the reduced model, which has the following relationship according to the OSF: 
\begin{equation}
    \mathbf{S}^T\Gamma = \mathbf{J}_r^T \mathbf{S}_r^T \Gamma_r + \mathbf{N}_r^T \Gamma_o,
\end{equation}
where $\mathbf{N}_r^T$ is the null-space projection matrix of $\mathbf{J}_r^T$ and $\mathbf{N}_r^T\Gamma_o$ represents the null-space torque vector that does not affect the reduced model space. 
Consequently, \eqref{eq:dynamics_reduce} can be considered a simplified representation of a system interacting with the external environment, and it can be utilized as equations of motion constraint for the reduced model in the iWBC method. 

The effect of the unconstrained chain on the external environment can be described using~\eqref{eq:dynamics_reduce}, following the simplification process. Since the simplified model cannot capture the internal motion of the unconstrained chain, it's dynamics must also be described.
By considering the unconstrained chain as an independent model on the base link's frame, it's dynamics is same with the rigid body dynamics on the fixed frame;
\begin{equation}\label{eq:dynamics_unconstrained}
    \mathbf{M}_{uc} \ddot{\mathbf{q}}_{uc} + \mathbf{b}_{uc} + \mathbf{g}_{uc} = \boldsymbol{\Gamma}_{uc},
\end{equation}
where $\mathbf{M}_{uc}$ is inertia matrix of the unconstrained chain, $\mathbf{b}_{uc}$ is the Coriolis and centrifugal, and $\mathbf{g}_{uc}$ is the gravity vector of unconstrained chain.

Through the process described in this subsection, the whole-body dynamics~\eqref{eq:dynamics} can be divided into~\eqref{eq:dynamics_reduce} and~\eqref{eq:dynamics_unconstrained}, thereby dividing the iWBC problem as well.

\subsection{Motion Control Tasks via Reduced Space }\label{subsection:taskspaceconstraint}
In the proposed method, task constraints are divided into two categories: those related to the reduced model joint space and those within the unconstrained chain.

\subsubsection{Task Constraints in the Reduced Model Space}
Tasks related to the robot's COM, and the tasks existing within the constraint chain fall into the first category. This also includes control of the upper body COM or the floating base.
The relationship between the reference velocity in a specific task space and the joint velocities is given by:
\begin{equation}\label{eq:taskspace}
    \mathbf{J}_{task}\mathbf{\dot{q}} = \mathbf{\dot{x}}^{ref}_{task},
\end{equation}
where $\mathbf{J}_{task}\in \mathbb{R}^{m\times {(n+6)}}$ is the task space Jacobian matrix,  $\mathbf{\dot{x}}^{ref}_{task}\in\mathbb{R}^{m}$ is the reference task space velocity vector, and $m$ is the DOF of the task space.

The null-space projection matrix $\mathbf{N}_{r}^{T}$ can be computed as follows:
\begin{equation}\label{eq:null}
\mathbf{N}_{r}^{T} = \mathbf{I} - \mathbf{J}_{r}^{T} {\bar{\mathbf{J}}_{r}}^T.
\end{equation}
Using the identity $\mathbf{I}=\mathbf{N}_{r}^{T} + \mathbf{J}_{r}^{T}{\bar{\mathbf{J}}_{r}}^T$ from~\eqref{eq:null}, the left-hand side of \eqref{eq:taskspace} can be represented as follows:
\begin{equation}\label{eq:nr}
    \mathbf{J}_{task}(\mathbf{N}_r +\mathbf{\bar{J}}_r \mathbf{J}_r)\mathbf{\dot{q}} = \mathbf{\dot{x}}^{ref}_{task}.
\end{equation}
Expanding and rearranging~\eqref{eq:nr} yields the following:
\begin{equation}\label{eq:reducedtaskspace}\mathbf{J}_{task}\mathbf{N}_r \mathbf{\dot{q}} + \mathbf{J}_{task}\mathbf{\bar{J}}_r \mathbf{\dot{q}}_r = \mathbf{\dot{x}}^{ref}_{task}.
\end{equation}
When the task space exists within the constraints chain, or involves controlling robot's centroidal space or the centroidal space of the unconstrained chain, $\mathbf{J}_{task}\mathbf{N}_r\mathbf{\dot{q}}=0 $ holds true. 
This is because the joint velocity of the unconstrained chain, which do not affect the centroidal space of the unconstrained chain, does not influence these types of task space.
Therefore, task spaces in these categories can describe \(\mathbf{\dot{x}}^{ref}_{task}\) through \(\mathbf{\dot{q}}_r\):
\begin{equation}
    \mathbf{J}_{task}\mathbf{\bar{J}}_r \mathbf{\dot{q}}_r = \mathbf{\dot{x}}^{ref}_{task}.
\end{equation}
These types of task constraints are considered in the first part of the iWBC problem, which is in the reduced model joint space. 
\subsubsection{Task Constraints in the Unconstrained Chain}
The second category includes task spaces within the unconstrained chain. For these tasks, the constraints are established as:
\begin{equation}
    {^b}\mathbf{J}_{task,uc}\dot{\mathbf{q}}_{uc}={^b}\dot{\mathbf{x}}^{ref}_{task,uc},
\end{equation}
where ${^b}\mathbf{J}_{task,uc}\in\mathbb{R}^{m\times n_{uc}}$ is the task Jacobian in the unconstrained chain relative to the reference frame of the robot base, and the ${^b}\dot{\mathbf{x}}^{ref}_{task,uc} \in \mathbb{R}^{m}$ is the reference task velocity from the reference frame of the robot base. 
The vector ${^b}\dot{\mathbf{x}}^{ref}_{task,uc}$ is computed based on the result of the first part of the problem, as it defines the movement of the robot base. 
Additionally, ${^b}\dot{\mathbf{x}}^{ref}_{task,uc}$ can also be defined by the task planner. 
\subsubsection{Reference Acceleration Constraint}
The reference acceleration constraint in the task space, $\ddot{\mathbf{x}}^{ref}_{task}$, can be constructed by differentiating the reference velocity \eqref{eq:reducedtaskspace} above.

\section{Problem Formulation}\label{section4:problemformulation}
\subsection{Lexicographic Quadratic Programming}
The inequality and equality constraints for iWBC, can be expressed in affine form corresponding to the equality and inequality constraints at each level in an optimization problem consisting of \( p \) priorities,
\begin{equation}
    \mathbf{A_i}\mathbf{x}+\mathbf{a_i}\leq0,\, \mathbf{B_i}\mathbf{x}+\mathbf{b_i}=0,\, i=1,2,...,p
\end{equation}
All linear constraints from each priority can be combined and expressed as follows:
\begin{align}
    \underbrace{
    \begin{bmatrix}
        \mathbf{A}_1\\
        \vdots \\
        \mathbf{A}_p
    \end{bmatrix}}_{\mathbf{A}} \mathbf{x}+
    \underbrace{
    \begin{bmatrix}
        \mathbf{a}_1\\
        \vdots\\
        \mathbf{a}_p
    \end{bmatrix}}_{\mathbf{a} } \leq 0 , \quad 
    \underbrace{
    \begin{bmatrix}
        \mathbf{B}_1\\
        \vdots \\
        \mathbf{B}_p
    \end{bmatrix}}_{\mathbf{B}}\mathbf{x}+
    \underbrace{
    \begin{bmatrix}
        \mathbf{b}_1\\
        \vdots\\
        \mathbf{b}_p
    \end{bmatrix} }_{\mathbf{b}}= 0
\end{align}
where $\mathbf{x} = [\,\mathbf{\ddot{q}}^T\:\mathbf{F}_c^T\:\boldsymbol{\Gamma}^T ]^T$, $\mathbf{A}\in\mathbb{R}^{m_a\times (2n+6+6c)}$, $\mathbf{a}\in\mathbb{R}^{m_a}$, $\mathbf{B}\in\mathbb{R}^{m_b\times (2n+6+6c)}$, $\mathbf{b}\in\mathbb{R}^{m_b}$. $m_a$ and $m_b$ are the total number of inequality and equality constraints of all priorities, respectively. $c$ is the number of contact links.

With the constraints formulation above, LQP \cite{escande2014hierarchical} is formulated as follows:
\begin{subequations}\label{eq:lqp}
\begin{equation}
    \underset{\mathbf{x},\mathbf{v},\mathbf{w}}{\text{min}}\,\lVert\mathbf{v} \rVert^2 + \lVert \mathbf{w}\rVert^2\tag{\ref{eq:lqp}}
\end{equation}
\begin{align}    
    \textrm{s.t.} \quad&\mathbf{V}(\mathbf{A} \mathbf{x} + \mathbf{a}) \leq \mathbf{v} \\
    & \mathbf{W}(\mathbf{B} \mathbf{x} + \mathbf{b}) = \mathbf{w}
\end{align}
\end{subequations}
where $\mathbf{V}\in\mathbb{R}^{m_a \times m_a}$, $\mathbf{W}\in\mathbb{R}^{m_b \times m_b}$ are weight matrices of inequality and equality constraints, respectively, and $\mathbf{v}\in\mathbb{R}^{m_a}$, $\mathbf{w}\in\mathbb{R}^{m_b}$ are slack variables for inequality and equality constraints, respectively.

This formulation of the LQP problem can be solved by hierarchical method \cite{herzog2016structured} or weighted method \cite{sathya2021weighted, sherali1982equivalent}.
The sequential method utilizes the null space basis matrix of the higher priority's equality constraint matrix to maintain the higher priority's constraint and uses the previously computed slack variables for the inequality constraints.
The weighted method performs optimization by appropriately utilizing weighting matrices to ensure that constraints at higher priority levels receive more focus.

\subsection{LQP Formulation with Reduced Dynamics}
The optimization problem utilizing the reduced dynamics is divided into two parts: LQP 1 and LQP 2.
LQP 1 uses the reduced dimension dynamics as constraints to account for interactions with the external environment. 
LQP 2 deals with the control of the unconstrained chain, based on the constraint of the simplified unconstrained chain determined by the solution of the LQP 1.

In both parts, we construct a cost function to calculate the movement that minimizes the robot's \textit{acceleration energy} \cite{bruyninckx2000gauss}, $\mathbf{E}_a$. The acceleration energy of the robot is defined as follows:
\begin{equation}
    \mathbf{E}_a = \frac{1}{2} \ddot{\mathbf{q}}^T\mathbf{M}\mathbf{\ddot{q}}.
\end{equation}

The first part of the optimization problem, LQP 1, utilizing the reduced dimension dynamics, is formulated similarly to \eqref{eq:lqp} as follows,

\begin{subequations}\label{eq:ocp2}
\begin{equation}
\min_{\ddot{\mathbf{q}}_r,\mathbf{F}_c,\mathbf{v}, \mathbf{w}}  \ddot{\mathbf{q}}_r^T \mathbf{M}_r \ddot{\mathbf{q}}_r + \lVert \mathbf{v}\rVert^2+ \lVert \mathbf{w}\rVert^2\tag{\ref{eq:ocp2}}
\end{equation}
\begin{align}
\textrm{s.t.} \quad&\mathbf{V}(\mathbf{A} \mathbf{x} + \mathbf{a}) \leq \mathbf{v} \label{eq:ocp2ineq}\\
    & \mathbf{W}(\mathbf{B} \mathbf{x} + \mathbf{b}) = \mathbf{w}\label{eq:ocp2eq}
\end{align}
\end{subequations}
Similar to how \eqref{eq:eomfloatingbase} is considered as a constraint in the original joint space, the reduced model dynamics \eqref{eq:dynamics_reduce} also consider only the top six rows of equations as constraints, resulting in the exclusion of torque from the decision variables.

Equations \eqref{eq:ocp2ineq} and \eqref{eq:ocp2eq} represent the constraints mentioned in Section \ref{section2:wholenbodycontrol}, formulated through reduced dynamics. For the task space, the constraints are constructed using the reduced dynamics, as discussed in subsection \ref{subsection:taskspaceconstraint}, which can only be composed through such task constraints.


The given solution $\ddot{\mathbf{q}}_r^*$ of LQP 1 \eqref{eq:ocp2} are utilized in the second problem, LQP 2. From the $\ddot{\mathbf{q}}_r^* = [(\ddot{\mathbf{q}}_{vc}^*)^T\,\, \ddot{\mathbf{q}}_{cc}^T\,\,({^b}\ddot{\mathbf{x}}^*_{G,uc})^T]^T$,  $\ddot{\mathbf{q}}_{vc}^*\in\mathbb{R}^6$, $\ddot{\mathbf{q}}_{cc}\in\mathbb{R}^{n_{cc}}$ denotes the floating base spatial acceleration, joint acceleration of contact chain, respectively, and ${^b}\ddot{\mathbf{x}}^*_{G,uc}\in\mathbb{R}^6$ denotes the spatial acceleration of the unconstrained chain's centroidal space in the base frame.

The second part of optimization problem to handle the unconstrained chain, LQP 2, is,
\begin{subequations}\label{eq:ocp3}
\begin{equation}
\min_{\ddot{\mathbf{q}}_{uc},\mathbf{v}_{uc},\mathbf{w}_{uc}} \ddot{\mathbf{q}}_{uc}^T \mathbf{M}_{uc} \ddot{\mathbf{q}}_{uc}+ \lVert \mathbf{v}_{uc}\rVert^2+\lVert \mathbf{w}_{uc}\rVert^2\tag{\ref{eq:ocp3}}
\end{equation}
\begin{align}
\textrm{s.t.}\,\, 
& \cdots \quad \text{(other eq. and ineq. constraints)\label{eq:ocp3_constraints}} 
\end{align}
\end{subequations}
where $\mathbf{v}_{uc}$ and $\mathbf{w}_{uc}$ are the constraint slack variables for unconstrained chain. The constraints of LQP 2 are also formulated with similar to~\eqref{eq:ocp2ineq}, and~\eqref{eq:ocp2eq}.

The important constraint in \eqref{eq:ocp3_constraints} is the equality constraint with the unconstrained chain's centroidal space, utilizing the solutions in LQP 1, ${^b}\ddot{\mathbf{x}}^*_{G,uc}$,
\begin{equation}\label{eq:nccomconstraint}
    {^b}\mathbf{J}_{G,uc} \ddot{\mathbf{q}}_{uc}+{^b}\dot{\mathbf{J}}_{G,uc} \dot{\mathbf{q}}_{uc} = {^b}\ddot{\mathbf{x}}_{G,uc}^*.
\end{equation}
With this equality constraint, the solution of LQP 2 does not affect the LQP 1, \eqref{eq:ocp2}. 

As mentioned in section~\ref{sub1}, overlapping joints included in both chain can exist. In such cases, the acceleration of these joints obtained from the solution in LQP 1 are also considered as constraints in LQP 2.

\begin{figure}[t]
\centering
\includegraphics[scale=1.0]{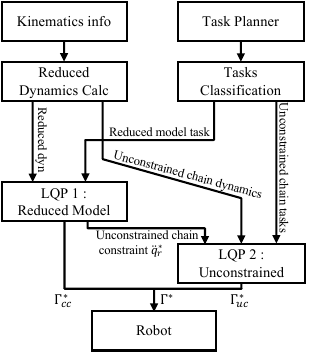}
\caption{The algorithm flow chart of the proposed method.}
\label{diagflow}
\end{figure}

For the task space constraint in the unconstrained chain, the desired acceleration of task space ,\({^b}\ddot{\mathbf{x}}^{ref}_{i,uc}\) is calculated from the reference task acceleration \(\ddot{\mathbf{x}}^{ref}_{i}\) and the acceleration of the floating base, $\ddot{\mathbf{q}}_{vc}$ obtained from the LQP 1. This allows for the computation of the desired acceleration \({^b}\ddot{\mathbf{x}}^{ref}_{i,uc}\) of the task relative to the robot base.
The overall flow of the calculation process can be seen in Fig.~\ref{diagflow}.


Depending on the priority of the unconstrained chain's tasks, which may take precedence over the centroidal space task, \eqref{eq:nccomconstraint} might need to be violated. This violation can be calculated in LQP 2 through the corresponding slack variables of constraint \eqref{eq:nccomconstraint}, $\mathbf{W}_{G,uc}^{-1}\mathbf{w}_{G,uc}^*$, and these violations can then be incorporated as an additional priority in LQP 1. Although the inclusion of this additional priority increases the computational load, the increase is not significant.

The constraint chain control torque $\boldsymbol{\Gamma}_{cc}^*\in\mathbb{R}^{n_{cc}}$ from the computation result of LQP 1 and the unconstrained chain control torque $\boldsymbol{\Gamma}_{uc}^*\in\mathbb{R}^{n_{uc}}$ from the LQP 2 are used to construct the control torque input of the robot $\boldsymbol{\Gamma}^* = [{\boldsymbol{\Gamma}_{cc}^*} ^T~{\boldsymbol{\Gamma}_{uc}^*}^T]^T$.

\section{Evaluation}\label{setion5:evaluation}

\subsection{System Setup}
 The proposed method was validated through the simulation. The robot used in the simulation, TOCABI, is human-sized (1.8m in height, 100kg in weight), 33-DOF torque-controlled humanoid robot. It is equipped with eight DOFs in each arm, six in each leg, three in the waist, and two in the head. The experiment was conducted on i7-10700 CPU, with the simulator MuJoco.
For the QP solver, one of the off-the-shelf solvers, OSQP, was used.
The robot's operating system is Ubuntu 20.04, with a real-time patch applied using Xenomai 3.2.1. 
Further detailed information about the robot and system configuration can be found in \cite{10113634}.

\begin{table}[t]\label{table1}
\renewcommand{\arraystretch}{1.1}
\caption{Task Configuration with Conventional Method}
\resizebox{0.48\textwidth}{!}{%
\begin{tabular}{cll}
\hline
\hline
Priority & Task & Size of constraint \\
\hline
1 & Dynamics constraint \eqref{eq:eomfloatingbase} & $6 \text{ eq}$\\
  & Toruqe limit & $2\times n \text{ ineq}$ \\
2 & Contact constraint & $6\times c \text{ eq}$\\
  & CoP \& friction cone & $10\times c \text{ ineq}$ \\
  & Joint acceleration limit & $2\times n \text{ ineq}$\\
3 & COM position & $3 \text{ eq}$ \\
  & Pelvis Orientation  & $3 \text{ eq}$\\
4 & End effector position \& orientation & $6\times k \text{ eq}$\\
\hline
\multicolumn{3}{l}{
decision variable size : $(n+6)+6\times c$}\\
\end{tabular}%
}
\label{tableTask}
\end{table}

\begin{table}[t]\label{table2}
\renewcommand{\arraystretch}{1.1}
\caption{Task Configuration with Proposed Method}
\resizebox{0.48\textwidth}{!}{%
\begin{tabular}{ccll}
\hline
\hline
 & Priority & Task & Size of constraint \\ \cline{2-4}
\multirow{7}{*}{LQP 1} & 1 & Dynamics constraint \eqref{eq:dynamics_reduce} & $6$ \text{ eq}\\
 & & Toruqe limit & $2\times n_r \text{ ineq}$ \\
 & 2 & Contact constraint & $6\times c \text{ eq}$\\
 & & CoP \& friction cone & $10\times c \text{ ineq}$ \\
 & & Joint acceleration limit & $2\times n_r \text{ ineq}$\\
 & 3 & COM position & $3 \text{ eq}$ \\
 &  & Pelvis Orientation  & $3 \text{ eq}$\\ \cline{2-4}
& \multicolumn{3}{l}{decision variable size : $(n_r+6)+6\times c$}\\
\hline
\multirow{4}{*}{LQP 2} & 1  & unconstrained chain centroidal acceleration & $6 \text{ eq}$ \\
           && Torque limit & $2\times n_{uc} \text{ ineq}$ \\
 & 2 & End effector position \& orientation & $6\times k \text{ eq}$\\
           && Joint acceleration limit & $2\times n_{uc} \text{ ineq}$ \\ \cline{2-4}
& \multicolumn{3}{l}{decision variable size : $n_{uc}$}\\
\hline
\hline
\end{tabular}%

}
\label{tableTask2}
\end{table}

\subsection{Result with Proposed Method}
The task for comparative experiments is configured as seen in Table~\ref{tableTask}. 
These tasks were constructed similarly to the tasks in \cite{herzog2016momentum}, for bipedal robot walking.
Table~\ref{tableTask} shows the various task hierarchies of LQP formulation configured using the original whole-body model.
Since the walking is consists of single contact and double contact scenarios, 
Table~\ref{tableTask2} shows how the tasks from Table~\ref{tableTask} are divided into two parts utilizing the proposed method.
The result measured the computation time in single support ($c=1$, $k=2$) and double support ($c=2$, $k=1$) scenarios occurring during walking.
The tasks configured in this manner were structured in an LQP formulation and computed through the sequential method. 

Table~\ref{tableComputenew} compares the average computation times of the conventional and proposed methods during double and single contact scenarios. The proposed method demonstrated a computation time decrease of 66.5\% in the single support scenario and 54.9\% in the double support scenario, indicating enhanced computational efficiency.



\begin{figure}
    \centering
    \includegraphics[width=0.46\textwidth]{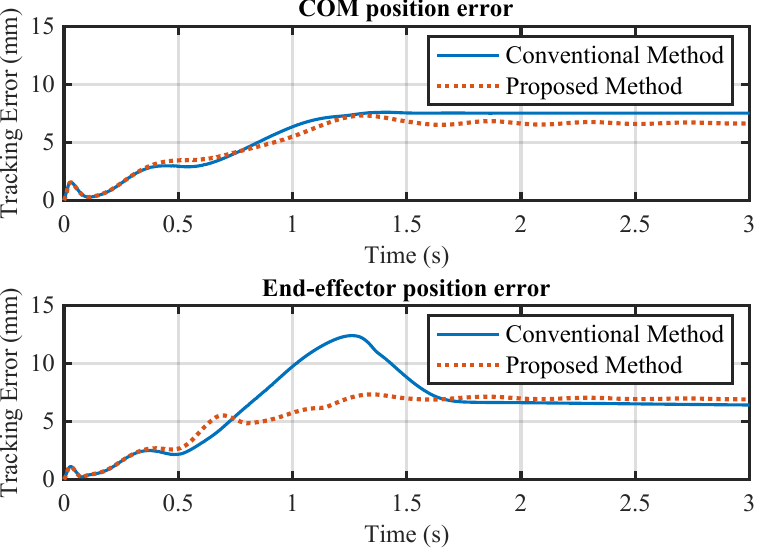}
    \caption{Tracking errors of COM and end-effector positions using conventional and proposed methods in the double support.}
    \label{fig:graph1-2}
\end{figure}
The trajectory tracking errors of the proposed and conventional methods were measured while simultaneously controlling the COM and the arm’s end-effector. As shown in Figure~\ref{fig:graph1-2}, both methods exhibited similar magnitudes of control error. Although task space control performance depends on the tuning of the weighting matrix—making exact comparisons challenging—both methods achieved comparable levels of control.
\begin{table}[ht]
    \centering
    \caption{Computation Time Results by Support types}
        \resizebox{0.48\textwidth}{!}{
    \begin{tabular}{llrrrrr}
        \hline
        \hline
        {Support Type} & {Method} & {Total AVG} & Total Max & {Dyn Calc} & {LQP 1} & LQP 2 \\
        \hline
        \multirow{2}{*}{Double} & Conventional & 2176 $\mu$s & 2473 $\mu$s & - & - & - \\
    
        & Proposed & 981 $\mu$s& 1203 $\mu$s& 34 $\mu$s& 579 $\mu$s& 273 $\mu$s\\
        \hline
        \multirow{2}{*}{Single} & Conventional & 1834 $\mu$s & 2119 $\mu$s& - & - & - \\
        
         & Proposed & 613 $\mu$s& 811 $\mu$s& 36 $\mu$s& 243 $\mu$s& 270 $\mu$s\\
        \hline
    \label{tableComputenew}
    \end{tabular}
    }
    \label{tab:support_times}
\end{table}

\subsection{Compuational Performance Analysis by DOF}
The proposed method was analyzed for its impact on the computational efficiency of LQP based on the DOF of the robot’s unconstrained chain. With both feet in contact, benchmarks examined how varying the unconstrained chain’s DOF affected computational efficiency. 
The task configuration for the comparison is similar to that in Tables~\ref{tableTask} and \ref{tableTask2}, except for the task within the unconstrained chain. Specifically, the arm’s end-effector is defined as the task space.

\begin{figure}
    \centering
    \includegraphics[width=0.45\textwidth]{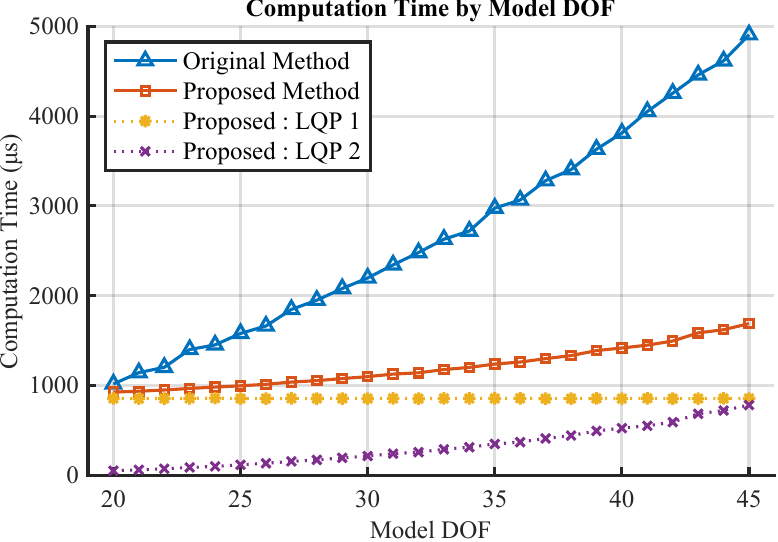}
    \caption{Average computation time comparison by DOF of the robot, and the relative computation time percentage. The computation time of proposed method is the sum of LQP 1 and LQP 2.}
    \label{fig:graph2}
\end{figure}

Benchmark tests were conducted on robot models with DOF ranging from 20 to 45, as shown in Figure~\ref{fig:graph2}. In the double contact scenario, the DOF of the unconstrained chain varies from 8 to 33. For a 20-DOF robot model, the proposed method required 571$\mu$s, while the conventional method took 754$\mu$s, indicating a 24.2\% reduction in computation time. This efficiency gain becomes more pronounced with an increase in the DOF of the unconstrained chain. When tested on a 45-DOF robot model, the proposed method took only 1,319$\mu$s, whereas the conventional method required 4,081$\mu$s, resulting in a 67.7\% decrease in computation time.

\section{CONCLUSIONS}\label{section6:conclusion}
This method employs reduced-dimension dynamics based on the constraints of the kinematic chain by partially projecting the unconstrained kinematic chain into its centroidal space, aiming to improve computational efficiency.
The proposed method offers a unique approach by dividing the model into two parts with lower DOF, thereby enhancing computational efficiency by handling lower-DOF models in each sequence of the problem.

This strategy distinguishes the proposed method from conventional approaches to iWBC by focusing on dynamics simplification, which allows for the application of conventional methods of iWBC.
The proposed method is integrated with LQP-based problems using the sequential method with off-the-shelf solvers. It is also available to use other solving method of LQP, or integrate the other dedicated solvers for the LQP.
 
When the proposed method was applied to the LQP formulation, which is commonly used in other studies, and computed using the sequential method, a 54.9 \% decrease in computation time was observed for the 33-DOF model at double support, and a 67.7\% decrease was observed for the 45-DOF model. This demonstrates a significant efficiency improvement with increasing DOF. The control performance was also found to be similar between the two methods, demonstrating that computational time can be reduced without compromising control performance.

For the future works, we would also like to develop this method for higher-level planners to tackle more complex computational challenges.

\section*{CONFLICT OF INTEREST}

The authors declare that there is no competing financial
interest or personal relationship that could have appeared
to influence the work reported in this paper.

\bibliographystyle{ieeetr}
\bibliography{ref}

\biography{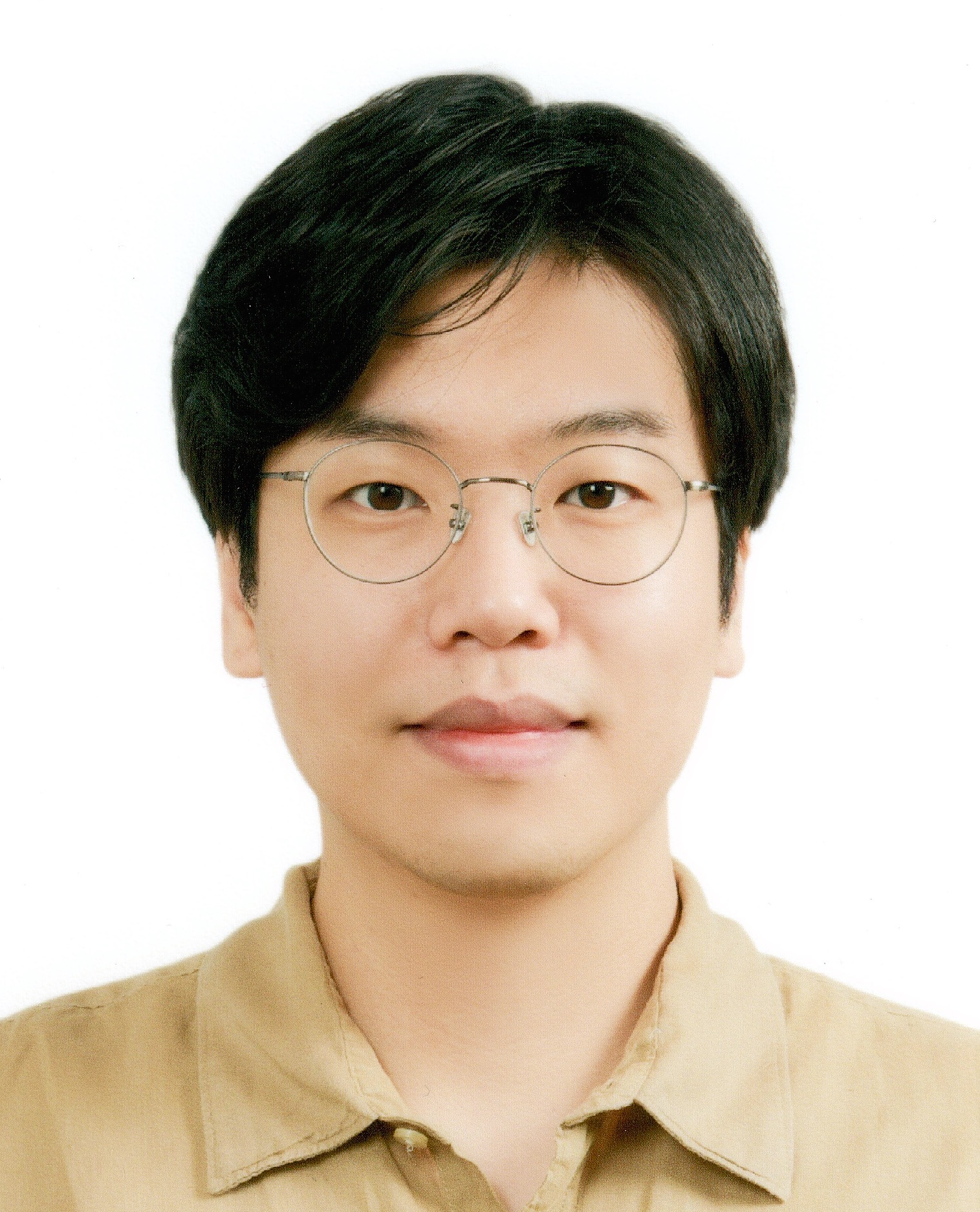}{JuneWhee Ahn}{received the B.S. degree in mechanical engineering from Sungkyunkwan University, Seoul, South Korea, in 2016. He is currently pursuing the Ph.D. degree with the Department of Intelligence and Information, Seoul National University, Seoul. His research interests include whole body control of a humanoid robot, optimal control, and control system framework.}

\biography{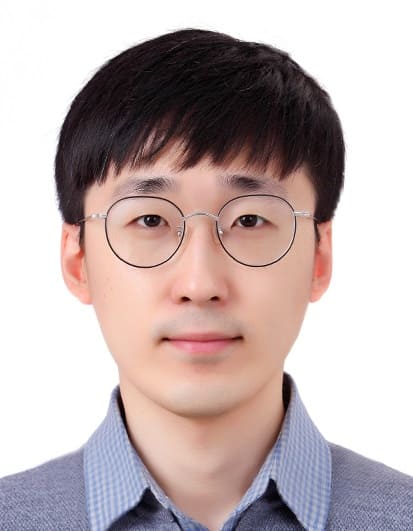}{Jaesug Jung}{received the B.S. degree in electrical and computer engineering from Seoul National University, South Korea, in 2015, and the Ph.D. degree in Department of Intelligence and Information, Seoul National University, in 2022. He worked as a  postdoctoral researcher at Technical University of Munich from 2022 to 2024. His research interests include torque-controlled robots, contact control, and whole-body control.}

\biography{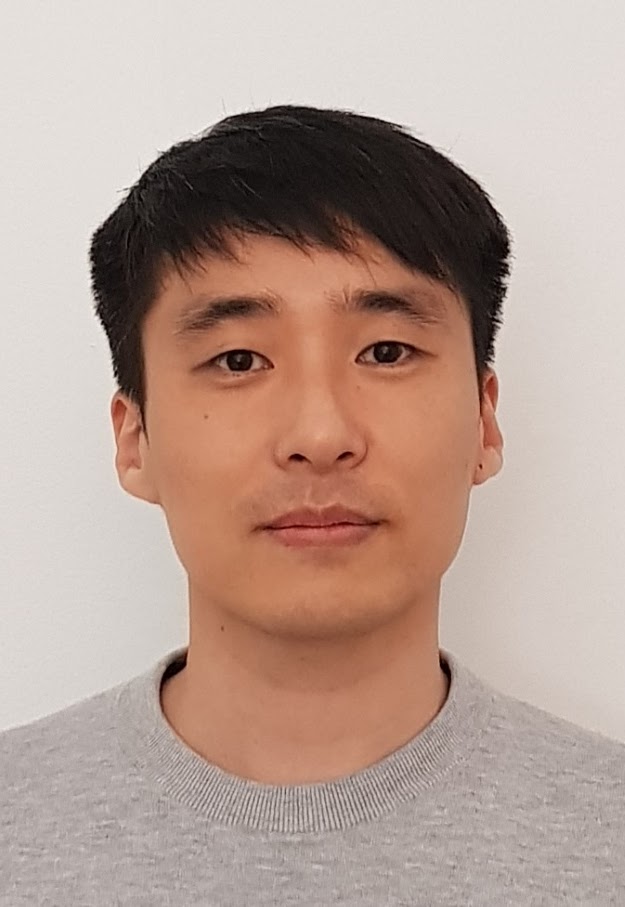}{Yisoo Lee}{Yisoo Lee received the B.S. and M.S. degrees in naval architecture and ocean engineering and the Ph.D. degree in intelligent convergence systems from the Department of Intelligent Convergence Systems from Seoul National University, Seoul, South Korea, in 2008, 2010, and 2017, respectively. 
From 2017 to 2018, he was a Postdoctoral Researcher with the Seoul National University. From 2018 to 2020, he was a Postdoc with the Istituto Italiano di Tecnologia (IIT), Genoa, Italy. He is currently a Senior Researcher with the Korea Institute of Science and Technology (KIST), Seoul, South Korea. His research interests include humanoid robots, robot locomotion, manipulation, optimal control, and machine learning for robot control.}

\biography{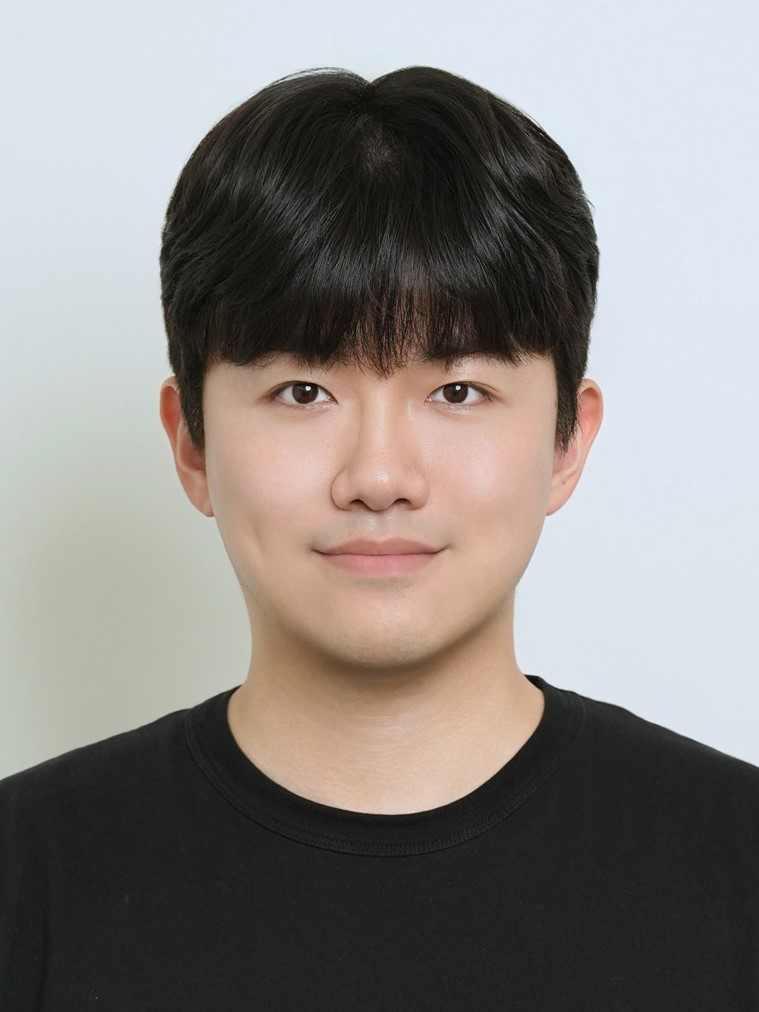}{Hokyun Lee}{received the B.E. degree in school of undergraduate studies from DGIST, Daegu, South Korea, in 2021. He is currently pursuing the Ph.D. degree with the Department of Intelligence and Information, Seoul National University, Seoul, South Korea. His research interests include whole-body dynamic control, reinforcement learning-based control, and humanoid robotics.}

\biography{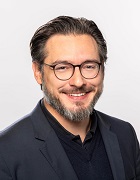}{Sami Haddadin}{received the Diploma.-Ing. degree in electrical engineering and the M.Sc. degree in computer science from Technical University of Munich (TUM), Munich, Germany, in 2005 and 2009, respectively, the Honors degree in technology management from Ludwig Maximilian University, Munich, Germany, and TUM, in 2007, and the Ph.D. degree in safety in robotics from RWTH Aachen University, Aachen, Germany, in 2011.

He is currently a Full professor and the Chair of Robotics and Systems Intelligence, Technical University of Munich and the founding Director of the Munich Institute of Robotics and Machine Intelligence. His research interests include physical human-robot interaction, nonlinear robot control, real-time motion planning, real-time task and reflex planning, robot learning, optimal control, human motor control, variable impedance actuation, and safety in robotics.}


\biography{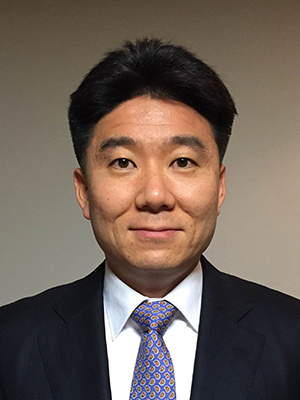}{Jaeheung Park}{received the B.S. and M.S. degrees in aerospace engineering from Seoul National University, South Korea, in 1995 and 1999, respectively, and the Ph.D. degree in aeronautics and astronautics from Stanford University, USA, in 2006. From 2006 to 2009, he was a Postdoctoral Researcher and later a Research Associate with the Stanford Artificial Intelligence Laboratory. From 2007 to 2008, he worked part-time with Hansen Medical Inc., a medical robotics
company, USA. Since 2009, he has been a Professor with the Department of Intelligent Convergence Systems, Seoul National University. His research interests include robot-environment interaction, contact force control, robust haptic teleoperation, multicontact control, whole-body dynamic control, biomechanics, and medical robotics.}

\clearafterbiography
\relax 

\end{document}